\documentclass[10pt,twocolumn,letterpaper]{article}

\usepackage{3dv}
\usepackage{times}
\usepackage{epsfig}
\usepackage{graphicx}
\usepackage{amsmath}
\usepackage{amssymb}
\usepackage{multirow}
\usepackage{amsthm}
\newtheorem{theorem}{Theorem}
\newtheorem{lemma}{Lemma}

\usepackage[pagebackref=true,breaklinks=true,letterpaper=true,colorlinks,bookmarks=false]{hyperref}

\threedvfinalcopy 

\def\threedvPaperID{154} 

\begin{document}

\title{Rethinking PointNet Embedding for Faster and Compact Model}

\author{Teppei Suzuki\\
Denso IT Laboratory\\
Tokyo, Japan\\
{\tt\small tsuzuki@d-itlab.co.jp}
\and
Keisuke Ozawa\\
Denso IT Laboratory\\
Tokyo, Japan\\
{\tt\small kozawa@d-itlab.co.jp}
\and
Yusuke Sekikawa\\
Denso IT Laboratory\\
Tokyo, Japan\\
{\tt\small ysekikawa@d-itlab.co.jp}
}

\maketitle

\begin{abstract}
PointNet, which is the widely used point-wise embedding method and known as a universal approximator for continuous set functions, can process one million points per second.
Nevertheless, real-time inference for the recent development of high-performing sensors is still challenging with existing neural network-based methods, including PointNet.
In ordinary cases, the embedding function of PointNet behaves like a soft-indicator function that is activated when the input points exist in a certain local region of the input space.
Leveraging this property, we reduce the computational costs of point-wise embedding by replacing the embedding function of PointNet with the soft-indicator function by Gaussian kernels.
Moreover, we show that the Gaussian kernels also satisfy the universal approximation theorem that PointNet satisfies.
In experiments, we verify that our model using the Gaussian kernels achieves comparable results to baseline methods, but with much fewer floating-point operations per sample up to 92\% reduction from PointNet.
\end{abstract}

\section{Introduction}

When developing robotics applications such as autonomous driving systems and simultaneous localization and mapping (SLAM), LiDAR is a useful sensor for capturing 3D geometry with point clouds in the scene, and methods that process the point clouds in real time are often required.
LiDAR captures million-order points per second, while the point clouds have intractability, such as unstructured and order permutation ambiguities.
Thus, the method requires processing over million-order points per second for real-time inference and invariance to permutation of points to deal with the point clouds.

Neural networks show remarkable results for point cloud recognition tasks, and some neural network-based methods transform the point clouds into tractable representations such as voxel and mesh~\cite{voxnet,gcnn,monet,fpnn,voxelnet,splatnet} to avoid the intractability.
However, these representations often cause information loss or require a large amount of memory~\cite{pvconv}.

Many methods of processing raw point clouds that satisfy the permutation invariant have been proposed, and they can avoid the information loss. The methods are roughly divided into two types: one is based on point-wise embedding~\cite{pointnet,deepsets,pointnet++}, and the other is based on a graph convolution~\cite{pointconv,kpconv,heat_kernel,monet,gcnn,spidercnn,shellnet}.
Typically, the graph convolution achieves better performance than the point-wise embedding methods because the graph convolution can capture local geometry.
However, because the graph convolution for point clouds requires the K-nearest neighbor search and random memory access to convolve points, it is often slower than point-wise embedding.
PointNet~\cite{pointnet}, which is a pioneer work for the point-wise embedding method, can process one million points per second on modern GPUs, but an advanced LiDAR sensor\footnote{https://velodynelidar.com/press-release/velodyne-lidar-debuts-alpha-prime-the-most-advanced-lidar-sensor-on-the-market/.} can obtain over four million points per second, so PointNet is still slow for that sensor.
The point-voxel CNN (PVCNN)~\cite{pvconv} tackled speedup and improvement of performance, and it achieved two times speedup over PointNet as well as slightly improved performance.
Developing faster methods is still important for real-time processing of advanced sensor data.

In this work, we propose a method that can reduce the computational cost of point cloud embedding, which dominates the entire computational cost of PointNet and PVCNN.
Our approach utilizes the property of the embedding function of PointNet.
The embedding function realized by a multi-layer perceptron (MLP) is known to behave like a soft indicator function~\cite{pointnet}, which is activated when the input points exist in a certain local region of the input space.
We explicitly define the embedding function as the Gaussian kernels, which work as the soft indicator function, and its floating-point operations per sample (FLOPs) is fewer than the MLP in PointNet.
Moreover, we provide a lemma that the Gaussian kernels also satisfy the universal approximation theorem for continuous set functions provided by Qi \textit{et al.}~\cite{pointnet}.
As a result, our method has the same representational capacity as PointNet, while dramatically reduces the computational costs.

Our contributions are summarized as follows: (i) We propose a new point-wise embedding model using Gaussian kernels to reduce the computational costs for embedding that dominates the entire computational time in PointNet~\cite{pointnet} and PVCNN~\cite{pvconv}, (ii) we show that the proposed method satisfies the universal approximation theorem, just as the original PointNet~\cite{pointnet}, and (iii) we show that the proposed model brings comparable performance to baseline methods~\cite{pointnet,luti,pvconv}, with much fewer floating-point operations per sample.
\section{Related Work}
Methods used for point clouds have been widely studied for many tasks, \textit{e.g.}, embedding~\cite{wave_kernel,fpfh}, recognition~\cite{heat_kernel,spin_img}, and registration~\cite{icp,sparse_icp,goicp}.
The point clouds are given by a set of points, and its order has no meaning.
Although neural networks are useful tools for point clouds, it is difficult for them to deal with the point clouds directly because outputs of the neural networks typically depend on the order of inputs.
Vinyals \textit{et al.}~\cite{ordermatter} naively used recurrent neural networks to deal with a set, but it is still not invariant to the permutation.

A straightforward solution for processing the point clouds with neural networks is to voxelize the point clouds and to then apply the volumetric convolution~\cite{c3d,voxnet,segcloud,efficient}, and many effective voxel-based methods have been proposed~\cite{voxnet,voxelnet,3dr2n2,octnet,3dunet,maturana20153d}.
As another approach, graph convolution methods have been recently studied~\cite{pointconv,gcnn,monet,acnn,edgeconv,kpconv,spidercnn} that utilize spatial locality of points.
However, voxel-based approaches have the trade-off between computational costs and accuracy~\cite{pvconv}, and graph convolution methods typically require high computational costs because of computing $K$-nearest neighbors and irregular memory access.

As the first work to process raw point clouds, PointNet~\cite{pointnet} impacted on the computer vision field.
PointNet solves the unordered problems by building the model as a symmetric function that is invariant to permutations of points.
Although it is difficult for PointNet to capture the local structure of point clouds unlike the volumetric convolutions and graph convolutions do, it can process over one million points per second on modern GPUs, and they show promising results for object classification, part segmentation, and semantic segmentation tasks.
This success suggested that point clouds could be represented just by the point-wise embedding and the maxpooling, and many methods were inspired by PointNet~\cite{pointpillars,frustum,pointnetlk,pointnet++,flownet3d,sgpn}.
These methods use PointNet as the backbone, and reduction of the computational costs of PointNet make these methods faster..
Zaheer \textit{et al.}~\cite{deepsets} also proposed permutation invariant models called DeepSets.
The design of DeepSets is similar to PointNet, and one can consider DeepSets as a generalized model of PointNet.

PointNet is a fast computation method for point clouds, but faster methods are required by the recent development of high-performing sensors.
PVCNN~\cite{pvconv} reduces the computational costs by reducing the number of embedding dimensions without hurting the performance by combining the point-wise embedding with the volumetric convolution.
As a result, PVCNN shows lower latency (2$\times$ faster) and slightly higher accuracy than PointNet.
LUTI-MLP~\cite{luti} also achieves lower latency (80$\times$ faster) than PointNet.
LUTI-MLP replaces the MLP used for point-wise embedding in PointNet with the lookup table to reduce the computational time of point-wise embedding.
Both methods show the effectiveness in terms of latency, but PVCNN has room for speedup because PVCNN uses the same embedding function as PointNet in part of its structure, and the embedding function of PointNet has redundancy as described in the next section.
On the other hand, the FLOPs of LUTI-MLP increase in an exponential order with respect to the dimension of the input point feature.

Under the same motivation as LUTI-MLP, we realize a computational time as fast as LUTI-MLP but with more scalability with respect to the input point feature dimension by using the Gaussian kernels.
We also provide a lemma that the Gaussian kernels satisfy the universal approximation theorem for the continuous set functions.
Moreover, we suggest two architectures for the segmentation tasks.
\begin{figure*}[t]
    \centering
    \includegraphics[clip,width=0.9\hsize]{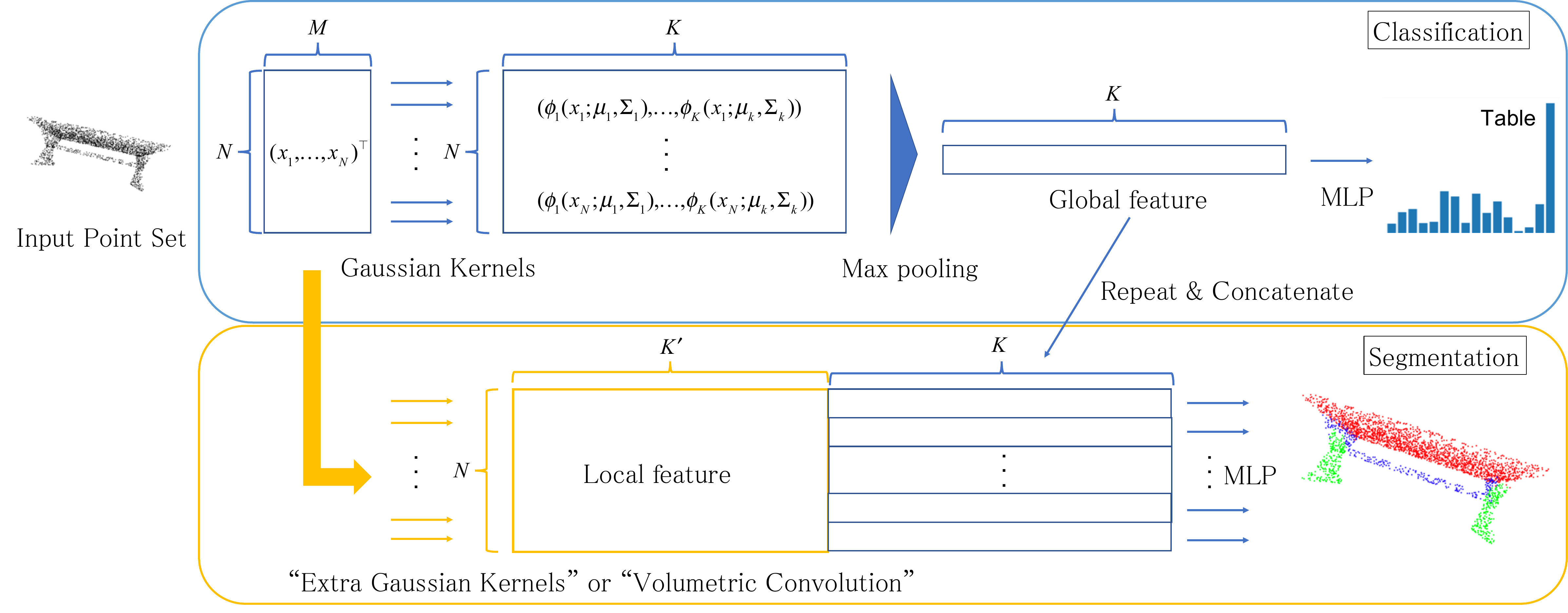}
    \caption{An overview of GPointNet. The classification model is the same as PointNet except for the embedding function. For segmentation, GPointNet has two options to obtain point-wise features: one is use of the extra Gaussian kernel, and the other is use of the volumetric convolution with voxelization, like PVCNN~\cite{pvconv}. The detailed architecture using the volumetric convolution can be found in Appendix \ref{app:exp_setting}}
    \label{fig:overview}
\end{figure*}

\section{PointNet Embedding as Gaussian Kernel}
The embedding framework of PointNet~\cite{pointnet} consists of point-wise MLP and channel-wise max pooling.
The point-wise MLP maps input point features (\textit{e.g.}, xyz-coordinate, color, and normal vector) into high-dimensional space.
The maxpool aggregation is an important concept of PointNet because it makes PointNet invariant to permutations of points.

The embedding function of PointNet is known to behave like a soft indicator function~\cite{pointnet}.
In other words, the value of each dimension of the maxpooled feature vector indicates whether the point exists in the region corresponding to the soft indicator function.
One can consider that PointNet samples points, which can represent the global shape of the point clouds, from input points.
Such a function can be realized with a simpler function than an MLP, and we explicitly define the embedding function as the soft indicator function to reduce computational costs.

Although there are many choices for the indicator function, we suggest use of Gaussian kernels as one reasonable choice because a Gaussian kernel can deal with multivariate data, its parameters are trainable in an end-to-end manner with backpropagation, the magnitude of the gradient with respect to the parameters is typically bound, and its computational costs are reasonable.

In the following sections, we review the universal approximation theorem provided by Qi \textit{et al.}~\cite{pointnet} and show that the Gaussian kernels satisfy the universal approximation theorem.

\subsection{Universal Approximation with Gaussian Kernel}

Let $\{\boldsymbol{x}_i\in\mathbb{I}^M\}_{i=1}^N\in\underbrace{\mathbb{I}^M\times\dots\times\mathbb{I}^M}_{N};\mathbb{I}=[a,b]$ be an input point set with $M$ dimensional features, where $a\in\mathbb{R}$ and $b\in\mathbb{R}$ are any scalar values, which means that we assume that the point features are normalized into a certain range, $[a,b]$.
First, PointNet embeds each point into a $K$ dimensional real space, $\{h_k(\boldsymbol{x}_i)\}_{k=1}^K$, with an MLP, $h_k:\mathbb{I}^{M}\rightarrow\mathbb{R}$.
Next, PointNet aggregates feature vectors with maxpooling, $\mathrm{MAX}(\{h_k(\boldsymbol{x}_i)\})=\{\max_i h_k(\boldsymbol{x}_i)\}_{k=1}^K;\ \mathrm{MAX}:\mathbb{R}^K\times\dots\times\mathbb{R}^K\rightarrow\mathbb{R}^K$, and obtains a $K$ dimensional vector.
Finally, PointNet classifies input points by the classifier $\gamma:\mathbb{R}^K\rightarrow\mathbb{R}$.
Because of the symmetric function $\mathrm{MAX}(\cdot)$, PointNet is invariant to permutations of input points.

We define $f:\mathbb{R}^M\times\dots\times\mathbb{R}^M\rightarrow\mathbb{R}$, satisfying a following condition as a continuous set function with respect to the Hausdorff distance $d_H(\cdot,\cdot)$ at $\mathcal{X}\in\mathbb{R}^M\times\dots\times\mathbb{R}^M$:
\begin{align}
    \nonumber
    \forall\epsilon>0,\exists\delta>0,\text{ such that for any }\mathcal{X}'\in\mathbb{R}^M\times\dots\times\mathbb{R}^M,\\ \nonumber
    d_H(\mathcal{X},\mathcal{X}')<\delta\Rightarrow|f(\mathcal{X})-f(\mathcal{X}')|<\epsilon,
\end{align}
and then Qi \textit{et al.}~\cite{pointnet} provides the following theorem:
\begin{theorem}
\label{theorem}
    Suppose $f:\mathbb{I}^M\times\dots\times\mathbb{I}^M\rightarrow\mathbb{R}$ is a continuous set function w.r.t Hausdorff distance $d_H(\cdot,\cdot)$. $\forall\epsilon>0$, $\exists$ a continuous function $h_k$ and a symmetric function $g(\boldsymbol{x}_1,\dots,\boldsymbol{x}_N)=\gamma\circ\mathrm{MAX}$, such that for any $\mathcal{X}=\{\boldsymbol{x}_i\in\mathbb{I}^M\}$,
    \begin{align}
        \nonumber
        \left| f(\mathcal{X})-\gamma\left(\mathrm{MAX}(\{h_k(\boldsymbol{x}_i)\}_{k=1}^K)\right)\right|<\epsilon,
    \end{align}
    where $\gamma(\cdot)$ is a continuous function.
\end{theorem}

Theorem \ref{theorem} indicates that PointNet can approximate the continuous set functions with any $\epsilon$ if $K$ is sufficiently large because $h_k$ and $\gamma$ correspond to the point-wise embedding MLP and the classification MLP of PointNet, respectively, and the MLP is known as the universal approximator.
However, Qi \textit{et al.}~\cite{pointnet} define $h_k(\cdot)$ as a soft indicator function in the proof, and in fact, they empirically showed the embedding MLP behaves like the soft indicator function as a training result.
Nevertheless, PointNet utilizes the MLP as $h_k(\cdot)$, and it requires hundred million floating-point operations.
Therefore, we introduce Gaussian kernel, $\phi_k:\mathbb{R}^M\rightarrow(0,1]$, which works as the soft indicator function, and its computational cost is reasonable than the MLP.

The introduced Gaussian kernel is defined as follows:
\begin{align}
    \label{eq:gi}
    \phi_k(\boldsymbol{x}_i)=\exp\left(-\left((\boldsymbol{x}_i-\boldsymbol{\mu}_k)^\top\mathbf{\Sigma}_k(\boldsymbol{x}_i-\boldsymbol{\mu}_k)\right)\right),
\end{align}
where $\boldsymbol{\mu}_k\in\mathbb{R}^M$ and $\mathbf{\Sigma}_k\in\mathbb{R}^{M\times M}$ denote a mean vector and an inverse covariance matrix, respectively.
We define the inverse covariance matrix as the positive semi-definite matrix.
Then, we provide the following lemma:
\begin{lemma}
\label{lemma}
Suppose $f:\mathbb{I}^M\times\dots\times\mathbb{I}^M\rightarrow\mathbb{R}$ is a continuous set function w.r.t Hausdorff distance $d_H(\cdot,\cdot)$, and let $\phi(\cdot)$ be the Gaussian kernel.
Then, $\forall\epsilon>0$, a symmetric function $g(\boldsymbol{x}_1,\dots,\boldsymbol{x}_N)=\gamma\circ\mathrm{MAX}$ exists such that for any $\mathcal{X}=\{\boldsymbol{x}_i\in\mathbb{I}^M\}$,
\begin{align}
    \nonumber
    \left|f(\mathcal{X})-\gamma\left(\mathrm{MAX}\left(\{\phi_k(\boldsymbol{x}_i)\}_{k=1}^K\right)\right)\right|<\epsilon.
\end{align}
\end{lemma}
\begin{proof}
To simplify the proof, we assume that the point feature dimension $M$ is 1, and $\mathbb{I}=[0,1]$, but it is easily generalized.

We define the mean vectors of the $k$-th Gaussian kernel as $\mu_k=\frac{1}{2K}+\frac{k-1}{K}$, and the covariance as an identity matrix without loss of generality.
Obviously, a pair $(i,k)$ exists such that $\phi_k(x_i)\geq\exp(-(2K)^{-2})$.

We define $\tau:\mathbb{R}^K\rightarrow\mathbb{I}\times\dots\times\mathbb{I}$ as $\tau(y)=\left\{\mu_k+\sqrt{-\log y};y\geq\exp(-(2K)^{-2})\right\}$.
Note that $\mu_k+\sqrt{-\log y}$ is a (partial) inverse mapping of the $k$-th Gaussian kernel, $\phi_k(x)=y;x\geq\mu_k\Rightarrow\tau(y)=\phi_k^{-1}(y)$. 
Let $\boldsymbol{v}=\mathrm{MAX}(\{\phi_k(\boldsymbol{x}_i)\}_{k=1}^K);\boldsymbol{v}\in\mathbb{R}^K$, and we define $\{\tau(\boldsymbol{v}_k)\}_{k=1}^K$ as $\hat{\mathcal{X}}$.
Then, because $\sup_{x\in\mathcal{X}}\inf_{\hat{x}\in\hat{\mathcal{X}}}d(x,\hat{x})\leq\frac{1}{2K}$ and $\sup_{\hat{x}\in\hat{\mathcal{X}}}\inf_{x\in\mathcal{X}}d(\hat{x},x)\leq\frac{1}{2K}$, $d_H(\mathcal{X},\hat{\mathcal{X}})\leq\frac{1}{2K}$.
From the definition of a continuous set function w.r.t Hausdorff distance, $|f(\mathcal{X})-f(\hat{\mathcal{X}})|<\epsilon$.
Therefore, considering $\gamma:\mathbb{R}^K\rightarrow\mathbb{R}$ as $\gamma=f\circ\tau$, $\left|f(\mathcal{X})-\gamma\left(\mathrm{MAX}\left(\{\phi_k(\boldsymbol{x}_i)\}_{k=1}^K\right)\right)\right|<\epsilon$.
Moreover, because $\mathrm{MAX}(\cdot)$ is a symmetric function, $\gamma\circ\mathrm{MAX}$ is also a symmetric function.
\end{proof}
The proof is the same as the original proof provided by Qi \textit{et al.}~\cite{pointnet} except for using the Gaussian kernels and modifying the representation of $\tau$.
Lemma \ref{lemma} indicates that the Gaussian kernels are one of the continuous functions satisfying Theorem \ref{theorem}.

\subsection{Gaussian PointNet}
We refer to PointNet~\cite{pointnet} using the Gaussian kernels for the embedding as GPointNet.
GPointNet has advantages with respect to model size and computational costs compared with other point-wise embedding methods such as PointNet, LUTI-MLP~\cite{luti}, and PVCNN~\cite{pvconv}.
Moreover, its FLOPs is $\mathcal{O}(M^2)$, which is reasonable than that of LUTI-MLP.

The trainable parameters of the Gaussian kernel are mean vectors, $\{\boldsymbol{\mu}_k\in\mathbb{R}^M\}_{k=1}^K$, and inverse covariance matrices, $\{\mathbf{\Sigma}_k\in\mathbb{R}^{M\times M}\}_{k=1}^K$, and these parameters can be trained in an end-to-end manner through backpropagation because the Gaussian kernel is differentiable with respect to the parameters.
We define the inverse covariance matrix as the positive semi-definite matrix, and to ensure it, we hold it as Cholesky factorized representation, $\mathbf{\Sigma}_k=\mathbf{LL}^\top$, where $\mathbf{L}\in\mathbb{R}^{M\times M}$ is the triangular matrix.
This parameterization not only ensures positive semi-definiteness but also decreases the number of parameters from $M^2$ to $\frac{1}{2}M(M+1)$.
Therefore, the number of trainable parameters of the Gaussian kernel is given by $K\cdot\left(\frac{1}{2}M(M+1)+M\right)$.
The embedding function of the simplest PointNet has $<150$K parameters, and the Gaussian kernels have only $9.2$K parameters with the same setting as PointNet.
Moreover, the Gaussian kernels also require fewer floating-point operations than PointNet.
We provide the detailed analysis in Section \ref{sec:complexity}.

The difference between PointNet and GPointNet is the soft-indicator shapes that they can represent.
PointNet allows various shapes as the soft indicator function because of the property of the neural network as the universal approximator.
On the other hand, GPointNet allows an ellipse as the shape of the indicator.

Therefore, we suggest GMPointNet, which is a Gaussian mixture version of GPointNet, to mitigate the limitation of the ellipse and increase the variety.
GMPointNet utilizes Gaussian mixture kernels instead of the Gaussian kernels, and the Gaussian mixture kernel is defined as follows:
\begin{align}
    f_k(x_i)=\sum_{l=1}^L\alpha_k^{(l)}\exp\left(-\left((\boldsymbol{x}_i-\boldsymbol{\mu}^{(l)}_k)^\top\mathbf{\Sigma}^{(l)}_k(\boldsymbol{x}_i-\boldsymbol{\mu}^{(l)}_k)\right)\right),
\end{align}
where $\{\alpha_k^{(l)}\in\mathbb{R}_+\}_{l=1}^L;\ \sum_{l=1}^L\alpha_k^{(l)}=1$ is a trainable mixture of coefficients; $L$ is the number of mixtures and is a hyperparameter; $\{\boldsymbol{\mu}_k^{(l)}\in\mathbb{R}^M\}_{l=1}^L$ and $\{\mathbf{\Sigma}_k^{(l)}\in\mathbb{R}^{M\times M}\}_{l=1}^L$ are a set of mean vectors and covariance matrices, respectively.
Practically, to satisfy $\sum_{l=1}^L\alpha_k^{(l)}=1$, $\alpha_k^{(l)}$ is given through the softmax function, $\alpha^{(l)}_k=\frac{\exp\left(\hat{\alpha}^{(l)}_k\right)}{\sum_l\exp\left(\hat{\alpha}^{(l)}_k\right)};\hat{\alpha}^{(l)}_k\in\mathbb{R}$.
The number of trainable parameters with the Gaussian mixture kernel is $L\cdot K\cdot\left(\frac{1}{2}M(M+1)+M+1\right)$.

Although GMPointNet is a straightforward extension of GPointNet and can represent more shapes than GPointNet, GMPointNet is not always better than GPointNet in practice, especially if $K$ is sufficiently large.
The fact indicates that the shapes are not necessarily important to recognize the point clouds.
We provide a detailed analysis in Section \ref{exp:ablation}.

\subsection{Implementation Detail}
GPointNet and GMPointNet are implemented by replacing the point-wise embedding MLP in PointNet with the Gaussian kernels and the Gaussian mixture kernels.
However, PointNet utilizes an intermediate feature vector for segmentation tasks, and our GPointNet does not have such an intermediate representation.
Therefore, we suggest two options for segmentation tasks: one is the use of extra Gaussian kernels, and the other is the use of volumetric convolution with voxelization.
The former uses the global feature and the point-wise feature, which correspond to the concept of PointNet, and the latter uses the global feature and the local geometric feature, which correspond to the concept of the PVCNN.
The overview is shown in Fig. \ref{fig:overview}, and the detailed architecture (\textit{e.g.}, the number of layers and the layer parameters) is found in Appendix \ref{app:exp_setting}.

To maximize performance, PointNet uses a transformation network (TNet) that aligns input points, and GPointNet can also use the TNet.
Moreover, we can also reduce the computational costs of the TNet by replacing the MLP with the Gaussian kernel because the TNet consists of the same architecture as PointNet.
Therefore, the TNet for GPointNet uses the Gaussian kernel in our experiments.
Note that Qi \textit{et al.}~\cite{pointnet} suggest the use of TNet for both input and intermediate representation.
However, if we try to use the intermediate TNet for GPointNet, there is no advantage for model size and speed (the complexity analysis is in Section \ref{sec:complexity}).
Thus, GPointNet uses the TNet only for the input vector, but we show that GPointNet with the input TNet achieves comparable results to PointNet with both TNets.

To reduce the dependence of training results on initial values, the covariance matrices are initialized identity matrices, and the mean vectors are initialized centroid coordinates of a regular grid in the input space.
Surplus mean vectors are randomly initialized.
For example, when $K=1024$ and $M=3$, 1000 mean vectors are initialized as the centroid of a $10\times10\times10$ grid in $\mathbb{I}^3$, and the remaining 24 mean vectors are randomly sampled from $\mathbb{I}^3$.

\section{Experiments}
First, we study the best practice of GPointNet in an ablation study.
Next, to verify that the Gaussian kernel shows a comparable performance to the MLP and other fast embedding methods, we compare GPointNet with PointNet~\cite{pointnet}, PVCNN~\cite{pvconv}, and LUTI-MLP~\cite{luti} with three types of standard tasks for the point clouds, \textit{i.e.}, object classification, part segmentation, and semantic segmentation tasks.
Finally, to show the effectiveness of our GPointNet with respect to size and speed, we evaluate floating-point operations per sample and the number of trainable parameters for GPointNet and other baseline methods.
There are several choices as the model architecture for PVCNN~\cite{pvconv}, and we choose PVCNN ($0.25\times C$) and PVCNN ($0.125\times C$) for the part segmentation and semantic segmentation, respectively, because we focus on computational costs in this work, and PVCNN ($0.25\times C$) and PVCNN ($0.125\times C$) are the fastest of the models provided by Liu \textit{et al.}~\cite{pvconv} for these tasks.
For comparison of the segmentation tasks, we evaluate two models, GPointNet with the extra Gaussian kernel and with the volumetric convolution.
We refer to the former as GPN w/Gaussian, and the latter as GPN w/Conv.
The detailed architectures of all models can be found in Appendix \ref{app:exp_setting}.

We use ModelNet40~\cite{modelnet} for the classification task, ShapeNet~\cite{shapenet} for the part segmentation task, and the Stanford 3D semantic parsing dataset (S3DIS)~\cite{indoor} for the semantic segmentation task.
We feed the XYZ-coordinate into the model for the object classification and the part segmentation, and feed the 9-dimensional vector of XYZ, RGB, and normalized location into the model for the semantic segmentation.
The input vectors are normalized into $[-1,1]^M$.

As comparison metrics, we use accuracy for the object classification, intersection over union (IoU) for the part segmentation, and both for the semantic segmentation.
Note that IoU for each category of the part segmentation is calculated as the average of IoUs over part classes of each category, and mIoU is calculated as the average over all parts.

The model parameters are shown in Table \ref{tab:param}.
\begin{table}[t]
    \centering
    \caption{The parameters for experiments. The input points are randomly sampled from original points. $M$, $N$, $K$, and $K^\prime$ correspond to the parameters in Fig. \ref{fig:overview}}
    \begin{tabular}{c|c|c|c}
    \hline
                            & ModelNet40  & ShapeNet & S3DIS \\ \hline
    $M$    & 3          & 3        & 9 \\
    $N$   & 1024      & 2048    & 4096 \\
    $K$      & 1024      & 2048    & 1024 \\
    $K^\prime$ & N/A & 832    & N/A \\ \hline
    \end{tabular}
    \label{tab:param}
\end{table}
All the settings of our experiments followed PointNet's experiments~\cite{pointnet} and the author's implementation.\footnote{https://github.com/charlesq34/pointnet.}
Note that the embedding dimension for PVCNN is reduced to 25$\%$ for ShapeNet and to 12.5$\%$ for S3DIS, and we also reduce the dimension of GPN w/Conv with the same ratio as with PVCNN.
We train the models with Adam~\cite{adam} with default settings (\textit{i.e.}, $\alpha=0.001$, $\beta_1=0.9$, $\beta_2=0.999$, and $\epsilon=10^{-8}$).
The learning rate is divided by 2 every 20 epochs.
The number of iterations is set to 250 for the object classification, 200 for the part segmentation, and 50 for the semantic segmentation.
We utilize a data-augmentation technique only for the object classification task.
We augment the point clouds by randomly rotating the object along the up-axis and jittering the position of each point by a Gaussian noise with zero mean and 0.01 standard deviation.
We use PyTorch~\cite{pytorch} for implementation of our methods and reproducing baseline methods.

\subsection{Ablation Study}
\label{exp:ablation}
To seek the best practice of our GPointNet and GMPointNet, we evaluate our models with several setups.
We divide the 9843 training data from ModelNet40 into 7000 data for training data and 2843 data for validation data, and we show the results evaluated on the validation data.

\subsubsection{Is GMPointNet Always Better than GPointNet?}
\begin{figure}
    \centering
    \includegraphics[width=0.9\hsize]{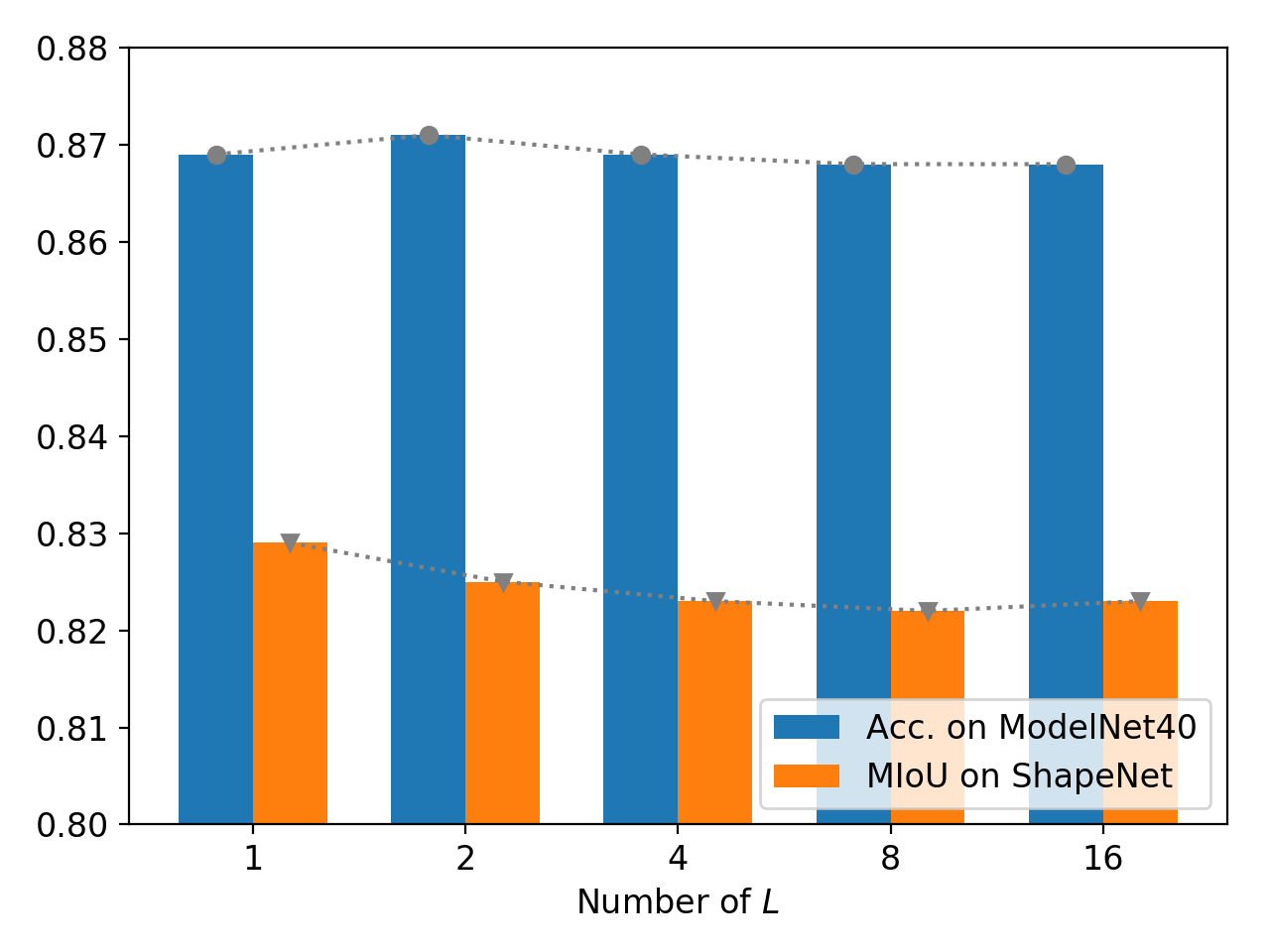}
    \caption{Accuracy and mean intersection over union (mIoU) of GPointNet ($L=1$) and GMPointNet with various mixture sizes $L$ evaluated with validation data from ModelNet40 and ShapeNet.}
    \label{fig:gi_vs_gmi}
\end{figure}

We evaluate GPointNet and GMPointNet with various mixture sizes $L$.
The results are shown in Fig. \ref{fig:gi_vs_gmi}.
GMPointNet with $L=2$ and $L=1$ (\textit{i.e.} GPointNet) is slightly better than the others for ModelNet40 and ShapeNet, respectively, but it is difficult to claim its advantage.
Moreover, changing the mixture size has no effect.
The result indicates that the shape of the soft indicator is not necessarily important for 3D object classification and part segmentation.

\begin{figure}
    \centering
    \includegraphics[width=0.9\hsize]{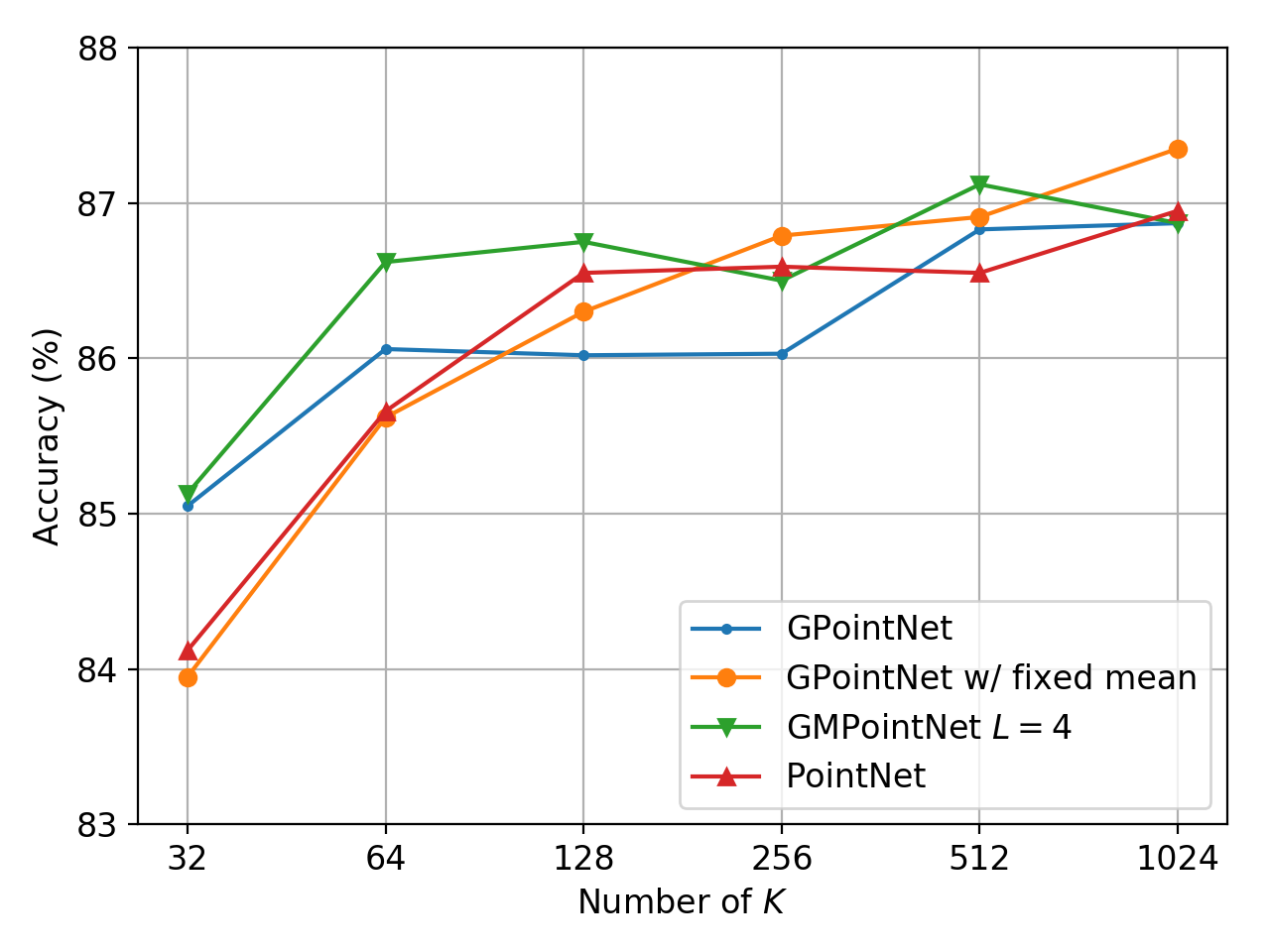}
    \caption{Accuracy (\%) on ModelNet40 with various embedding sizes.}
    \label{fig:K_vs_Acc}
\end{figure}

We assume that the results are achieved with sufficiently large $K$ for these tasks because the indicator function can completely describe input points if $K$ is infinite.
If so, the capacity of the soft indicator function in GMPointNet does not become an advantage.
Thus, we show the results with various $K$ in Fig. \ref{fig:K_vs_Acc}.
In fact, when $K=\{64,128\}$, GMPointNet achieves 0.5\% more accuracy than GPointNet, and GMPointNet is slightly better than GPointNet for all $K$.
However, the embedding of GMPointNet requires $L$ times more FLOPs and parameters than in GPointNet, and one should not waste the computational costs in most cases for the small improvement of accuracy.
When compared with PointNet, GPointNet and GMPointNet achieve better accuracy with a small $K$, \textit{i.e.}, 32 and 64.
We believe that the simplification of the embedding function helps to find a better embedding function, especially when $K$ is small.

For fair comparison of our method and PointNet, we set $K$ to 1024 in the following experiments so that GMPointNet has no advantage compared with GPointNet.
Therefore, we discuss only GPointNet in the following sections.

\subsubsection{Which Parameters Should Be Trained?}
If $K$ is sufficiently large, the mean vectors and the covariance matrices do not need to be trained because the indicator function can completely describe input points. Moreover, as seen in the proof of Lemma \ref{lemma}, the Gaussian kernel satisfies the universal approximation theorem with the initial values.
Thus, we verify whether we need to train the parameters.
\begin{table}[t]
    \centering
    \caption{Accuracy for ModelNet40 with various trainable parameters. Fixed mean and fixed covar indicate that the mean vectors and the covariance matrices, respectively, are fixed from the initial values.}
    \begin{tabular}{c|c|c|c}
        \hline
        & Fixed mean & Fixed covar & Accuracy (\%) \\ \hline
        GPointNet & & & 86.87 \\
        & \checkmark & & 87.35 \\
        & & \checkmark & 85.62\\
        & \checkmark & \checkmark & 86.02 \\ \hline
    \end{tabular}
    \label{tab:trainable}
\end{table}

As seen in Table \ref{tab:trainable}, the model with fixed mean vectors achieves the best accuracy.
However, this result depends on the size of $K$, and in fact, the mean vectors should be trained when $K$ is small (see Fig. \ref{fig:K_vs_Acc}).

\subsubsection{Dose TNet Enhance Model Performance?}
We verify the effect of the TNet for GPointNet, and the results are shown in Table \ref{tab:eff_tnet}.
\begin{table}[t]
    \centering
    \caption{Effect of TNet on ModelNet40.}
    \begin{tabular}{c|c|c}
        \hline
         & Input TNet  & Accuracy (\%) \\ \hline
         GPointNet &  & 86.87 \\
         & \checkmark  & 88.41 \\ \hline
         GPointNet &  & 87.35 \\
         w/fixed mean & \checkmark &  88.56\\ \hline
    \end{tabular}
    \label{tab:eff_tnet}
\end{table}
\begin{table}[t]
    \centering
    \caption{Results on ModelNet40~\cite{modelnet} with and without the input space TNet (I-TNet) and the feature space TNet (F-TNet).}
    \begin{tabular}{c|c|c|c}
        \hline
        & I-TNet & F-TNet & Accuracy (\%)\\ \hline
        PointNet~\cite{pointnet} & & & 87.07\\
        & \checkmark & & 88.09\\
        & \checkmark & \checkmark & 88.41\\ \hline
        LUTI-MLP~\cite{luti} & & & 85.98\\
        & \checkmark & & 88.33\\ \hline
        GPointNet & & & 87.40\\
        & \checkmark & & 88.53\\ \hline
    \end{tabular}
    \label{tab:modelnet}
\end{table}
\begin{table*}[t]
    \caption{IoUs (\%) on the ShapeNet part dataset~\cite{shapenet}. PN, LT, GPNG, PV, and GPNC denote PointNet~\cite{pointnet}, LUTI-MLP~\cite{luti}, GPN w/Gaussian, PVCNN (0.25$\times$ C)~\cite{pvconv}, and GPN w/Conv, respectively. The former three models do not use the local geometry features explicitly, while the latter models use them by the volumetric convolution.}
    \centering
    \scalebox{0.78}{
    \begin{tabular}{c|c|cccccccccccccccc}
        \hline
         & mIoU & aero & bag & cap & car & chair & \begin{tabular}{c}ear\\phone\end{tabular} & guitar & knife & lamp & laptop & motor & mug & pistol & rocket & \begin{tabular}{c}skate\\board\end{tabular} & table \\ \hline
         \#shapes & & 2690 & 76 & 55 & 898 & 3758 & 69 & 787 & 392 & 1547 & 451 & 202 & 184 & 283 & 66 & 152 & 5271 \\ \hline
        PN~\cite{pointnet} & 83.4 & 83.1 & 80.6 & 80.9 & 77.5 & 89.5 & 65.9 & 91.3 & 86.2 & 79.5 & 95.5 & 67.2 & 93.2 & 82.1 & 54.8 & 69.8 & 80.7 \\
        LT~\cite{luti}   & 81.6 & 81.7 & 72.2 & 80.1 & 72.6 & 87.5 & 61.9 & 89.8 & 83.4 & 78.0 & 94.4 & 62.7 & 92.8 & 78.8 & 55.1 & 72.0 & 79.8\\
        GPNG             & 83.1 & 82.5 & 74.2 & 72.6 & 72.7 & 89.1 & 71.3 & 90.7 & 85.4 & 78.8 & 93.8 & 63.4 & 92.7 & 80.6 & 51.4 & 72.1 & 81.8\\ \hline
        PV~\cite{pvconv} & 82.8 & 80.5 & 80.8 & 83.1 & 76.1 & 89.3 & 72.8 & 90.8 & 85.2 & 82.0 & 95.1 & 67.0 & 92.9 & 81.1 & 56.9 & 72.2 & 79.1 \\
        GPNC             & 82.4 & 79.8 & 72.3 & 82.3 & 73.9 & 89.3 & 68.3 & 89.1 & 85.4 & 80.2 & 95.0 & 55.2 & 91.4 & 78.0 & 47.0 & 71.0 & 81.5\\ \hline
    \end{tabular}}
    \label{tab:part_seg}
\end{table*}
Both GPointNet and its fixed mean version improve the accuracy by introducing the TNet.
Note that ``fixed mean'' fixes the mean vectors of both the TNet and the main module.

\subsection{Comparison of GPointNet with Baseline Methods}
In the following comparison, we fix the mean parameter of the Gaussian kernel, except for GPN w/Conv, and use the TNet only for the object classification tasks for a fair comparison with the corresponding baseline (\textit{i.e.}, GPointNet and GPN w/Gaussian vs. PointNet~\cite{pointnet} and LUTI-MLP~\cite{luti}, and GPN w/Conv vs. PVCNN~\cite{pvconv}).

\subsubsection{Object Classification.}
We compare our GPointNet with PointNet~\cite{pointnet} and LUTI-MLP~\cite{luti} on ModelNet40~\cite{modelnet}.
ModelNet40 has 12311 CAD models from 40 object categories.
The data are split into 9843 data for training data and 2468 data for the test data, and we evaluate the models on test data.

As seen in Table \ref{tab:modelnet}, GPointNet shows comparable or slightly better results than PointNet and LUTI-MLP.
Both GPointNet and LUTI-MLP with single TNet demonstrate the comparable accuracy to PointNet with two TNets.
In other words, PointNet requires the extra TNet to achieve reasonable results.
Thus, GPointNet and LUTI-MLP have an advantage for the computational costs in terms of their architectures.

\subsubsection{Part Segmentation.}
We evaluate the model performance with the part segmentation task on ShapeNet~\cite{shapenet}.
ShapeNet contains 16881 shapes from 16 categories, which are annotated with 50 part classes.
The shapes are split into 12137 data for training data, 1870 data for validation data, and 2874 data for test data.
We train the models with training data and evaluate them with test data.

We show the comparison results in Table \ref{tab:part_seg}.
Our GPointNet shows comparable results to PointNet and PVCNN, and outperforms LUTI-MLP.
Because of the trilinear interpolation of LUTI-MLP, LUTI-MLP has local linearity, and we consider that the linearity hurts the performance of LUTI-MLP for the part segmentation task.



\subsubsection{Semantic Segmentation.}
We evaluate the models with the semantic segmentation task on S3DIS~\cite{indoor} with a k-fold strategy used in PointNet.
S3DIS contains 3D scans from Matterport scanners in six areas, including 271 rooms.
Note that the input dimension $M$ is set to 9 for the semantic segmentation, and then, the complexity of LUTI-MLP is unreasonable.
Therefore, we do not evaluate LUTI-MLP because our GPU resources are limited.

\begin{table}[t]
    \centering
    \caption{Results with Stanford 3D semantic parsing dataset~\cite{indoor}. Mean intersection over union (IoU) is calculated as the average of IoU over 13 classes, and accuracy is calculated on points.}
    \begin{tabular}{c|c|c}
        \hline
         & Mean IoU (\%) & Overall acc. (\%)\\ \hline
        PointNet~\cite{pointnet} & 47.66 & 78.38 \\ 
        GPN w/Gaussian & 47.09 & 77.44 \\ \hline
        PVCNN ~\cite{pvconv} & 48.28 & 80.16 \\
        GPN w/Conv & 48.74 & 79.63 \\ \hline 
    \end{tabular}
    \label{tab:ss}
\end{table}
Our GPointNet achieves comparable results with the corresponding baselines.
PVCNN and GPN w/Conv achieve better mean IoUs and overall accuracy than PointNet and GPN w/Gaussian, with fewer embedding dimensions.
The results indicate that the local geometry is important for scene recognition.
PVCNN and GPN w/Conv also have the advantage in terms of the FLOPs per sample because of the reduction of the embedding dimension.
We show the detailed FLOPs/sample in Section \ref{sec:complexity}.


\begin{table}[t]
    \centering
    \caption{The number of parameters (\#param) and floating-point operations per sample (FLOPs/sample) for various embedding models. Those of PointNet are calculated with the model for ModelNet40. $M$, $K$, and $N$ denote input dimension, output dimension, and the number of input points, respectively. $D$ denotes a discretization parameter for LUTI-MLP. $E$ denotes FLOPs/sample of $\exp(\cdot)$ function. Note that the FLOPs/sample of PointNet ignores FLOPs/sample of batch normalization and ReLU activation because there are much fewer FLOPs/sample than the FLOPs/sample of MLP.}
    \scalebox{0.95}{
    \begin{tabular}{c|c|c}
    \hline
    \multirow{2}{*}{PN~\cite{pointnet}} & \#param & $64M+128K+$16384 \\ \cline{2-3} 
                      & FLOPs/sample & $N\cdot(128M+255K+$32448$)$ \\ \hline
    \multirow{2}{*}{LT~\cite{luti}} & \#param & $K\cdot D^M$ \\ \cline{2-3} 
                      & FLOPs/sample & $K\cdot N\cdot(2^M\cdot M+2^M-1)$ \\ \hline
    \multirow{2}{*}{Ours} & \#param & $K\cdot(\frac{1}{2}M\cdot(M+1)+M)$ \\ \cline{2-3} 
                      & FLOPs/sample & $K\cdot N\cdot(2M^2+2M-1+E)$ \\ \hline
    \end{tabular}
    }
    \label{tab:param_n_flops}
\end{table}
\begin{table*}[t]
    \centering
    \caption{
    The number of parameters (\#params) and floating-point operations per sample (FLOPs/sample) for various embedding models and common classifiers in the experiments.
    K and M stand for thousand and million, respectively.
    Note that the FLOPs/sample of GPointNet ignores $\exp(\cdot)$ FLOPs, because it depends on implementation.
    As an example for comparison, we show FLOPs/sample in parentheses if $\exp(\cdot)$ is calculated by the table lookup and the fourth-order Taylor approximation.
    }
    \label{tab:param_flops_modelnet}
    \scalebox{1}{
    \begin{tabular}{c|cc|cc|cc}
        \hline
           & \multicolumn{2}{c|}{Classification} & \multicolumn{2}{c|}{Part segmentation} & \multicolumn{2}{c}{Semantic segmentation}\\ \hline
           & \#param        & FLOPs/sample        & \#param       & FLOPs/sample     & \#param       & FLOPs/sample\\ \hline
PointNet~\cite{pointnet}   & 147.6K         & 301M                & 1.14M         & 4659M            & 148.0K & 1207M\\
LUTI-MLP~\cite{luti}   & 524.3K         & 32.5M               & 1.47M         & 183M             & 1.374E+11 & 2.147E+10\\
GPointNet & \textbf{9.2K}  & $>$\textbf{24.1}M(35.6M)   & - & - & - & -          \\ 
GPN w/ Gaussian & - & - & \textbf{0.03}M & $>$\textbf{136}M(201M) & \textbf{55.3}K & $>$\textbf{751}M(797M)\\ \hline 
Classifier & 665.6K         & 1.33M               & 0.85M         & 5611M            & 724.2K & 5630M\\ \hline\hline
PVCNN~\cite{pvconv}      & -            & -                   & 0.18M         & 1658M & 23.3K & 416M\\
GPN w/ Conv & -  & -                 & \textbf{0.11}M         & $>$\textbf{956}M(968M) & \textbf{15.8}K & $>$\textbf{293}M(299M)\\ \hline 
Classifier & -  & -                 & 0.09M         & 358M            &  14.3K &  116M \\\hline  
    \end{tabular}
    }
\end{table*}
\subsection{Complexity Analysis}
\label{sec:complexity}
First, we compare the point-wise embedding complexity of our GPointNet with that of PointNet and LUTI-MLP.
The number of parameters (\#param) and FLOPs per sample for various embedding models are summarized in Table \ref{tab:param_n_flops}.
Calculation of FLOPs/sample follows \cite{flops}, and the detail can be found in Appendix \ref{app:flops}.
Note that the constant values of PointNet in Table \ref{tab:param_n_flops} correspond to the parameters of intermediate layers that do not depend on the input and output dimensions.

\#param and FLOPs/sample of LUTI-MLP increase in an exponential order with respect to the discretization parameter and input dimension, while those of GPointNet increase in a squared order with respect to the input dimension.
Therefore, the complexity of LUTI-MLP may explode, depending on the input dimensions, but ours is somewhat robust to increasing input dimensions.
In fact, when considering the parameters of the semantic segmentation model (\textit{i.e.}, $M=9,\ K=1024,\ D=8, \text{and }N=4096$), LUTI-MLP has 1.4E+11 \#param and 2.1E+10 FLOPs/sample.
These complexities are much larger than those of PointNet, which are 6.1E+05 \#param and 1.2E+09 FLOPs/sample.
On the other hand, the complexity of GPointNet is 5.5E+04 \#param and $>$7.5E+07 FLOPs/sample.
Thus, in terms of scalability, GPointNet has an advantage compared with LUTI-MLP.

If GPointNet uses the TNet for intermediate feature vectors, we need two Gaussian kernels, where the first Gaussian kernel embeds the input points into the $K^\prime$ dimension, the TNet transforms the $K^\prime$ dimensional vectors, and the second Gaussian kernel embeds the $K^\prime$ dimensional vectors into the $K$ dimension.
Then, the second Gaussian kernel requires $K\cdot(\frac{1}{2}K^\prime\cdot(K^\prime+1)+K^\prime)$ \#param and $K\cdot N\cdot(2K^{\prime 2}+2K^\prime-1+E)$ FLOPs/sample.
For example, if $K^\prime=64$ and $K=1024$, GPointNet requires 2M \#param and $>$8723M FLOPs/sample.
There is no advantage for GPointNet, but essentially, GPointNet with only the input TNet shows comparable results to PointNet with both TNets. Also, in that case, GPointNet is still reasonable compared to PointNet in terms of computational complexity.

We show the number of parameters and FLOPs/sample for the models used in the experiments in Table \ref{tab:param_flops_modelnet}.
GPN w/Gaussian shows fewer \#param than LUTI-MLP and achieves comparable speedup.
Moreover, FLOPs/sample and \#param of LUTI-MLP explode for the semantic segmentation, while GPN w/Gaussian still achieves fewer \#param and FLOPs/sample than PointNet, as already described.
In terms of the entire computational cost, GPN w/ Gaussian for the semantic segmentation does not obtain a gain because the classifier dominates the entire computational cost.
However, the speedup for the segmentation tasks has already been achieved by the embedding dimension reduction in PVCNN, and GPN w/ Conv achieves further reduction of FLOPs/sample.
As a result, GPN w/ Conv achieves $<$94\% FLOPs/sample reduction from PointNet for the part segmentation, and $<$87\% FLOPs/sample reduction for semantic segmentation.


\section{Conclusion}
We proposed GPointNet, which is a model using a Gaussian kernel for point-wise embedding, and we showed that it was the universal approximator for continuous set functions.
GPointNet reduces the number of parameters and FLOPs per sample compared with PointNet and PVCNN, and it also demonstrates results comparable to baseline methods, with fewer FLOPs/sample, \textit{e.g.}, up to 92\% entire FLOPs/sample reduction from PointNet and up to 35\% entire FLOPs/sample reduction from PVCNN.

GPointNet reduces the computational costs of the embedding function.
In other words, GPointNet is a fast computational method for capturing the global feature.
The global feature obtained by PointNet is known to be useful for shape correspondence, model retrieval, and point cloud registration~\cite{pointnet,pointnetlk}.
If the global feature obtained by GPointNet is also useful for these tasks, our method provides the benefits of speedup for these tasks.
Thus, we will explore the effectiveness of the global feature obtained by GPointNet for other tasks in future work.

{\small
\bibliographystyle{ieee}
\bibliography{egbib}
}


\newpage
\begin{appendix}
\setcounter{figure}{0}
\renewcommand{\thefigure}{\Alph{section}\arabic{figure}}
\setcounter{equation}{0}
\renewcommand{\theequation}{\Alph{section}\arabic{equation}}

\section{Model Architecture}
\label{app:exp_setting}
\begin{figure}[t]
    \centering
    \begin{tabular}{c}
    \includegraphics[clip,width=1\hsize]{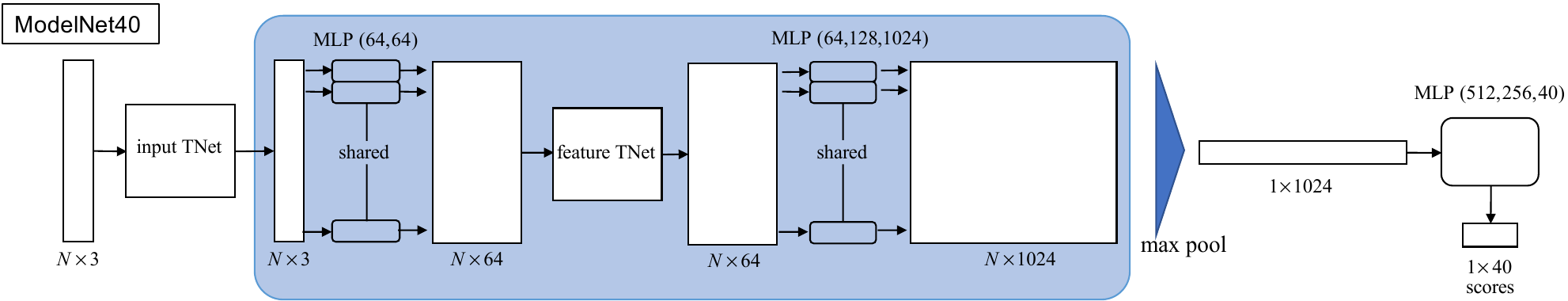}\\
    \includegraphics[clip,width=1\hsize]{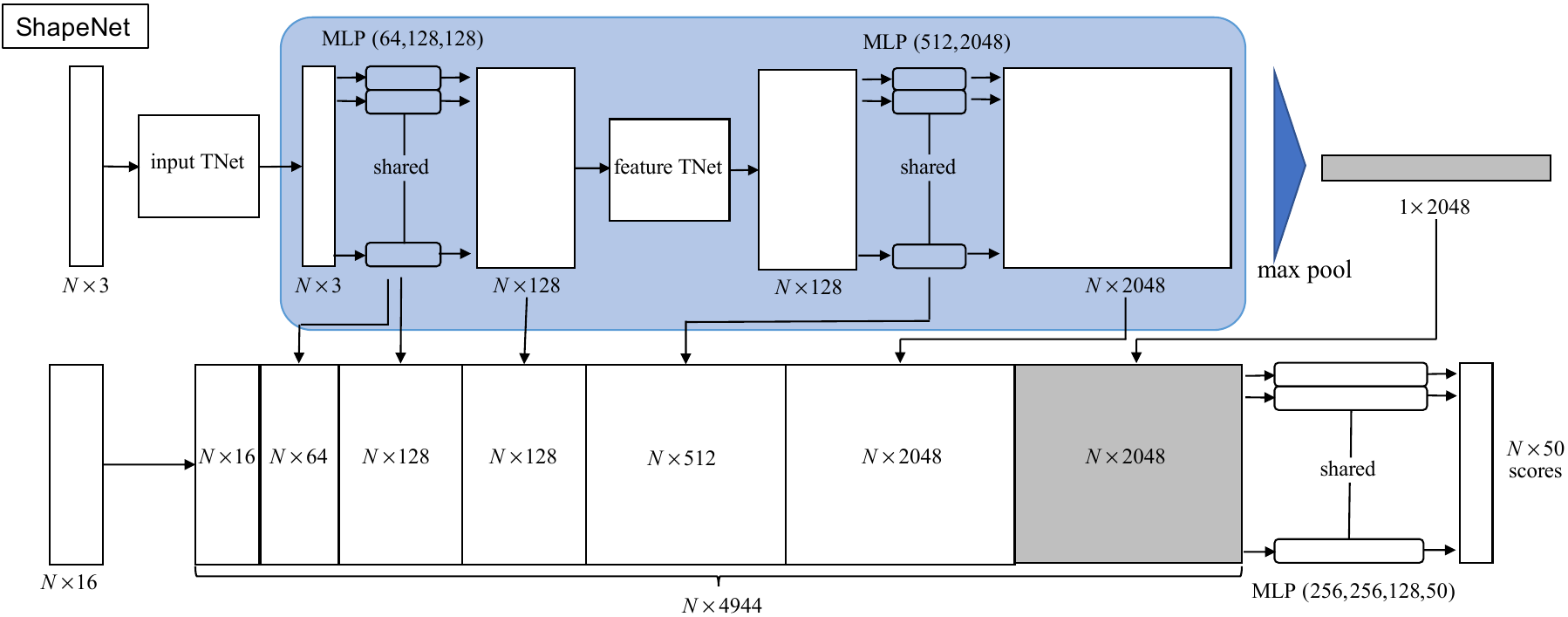}\\
    \includegraphics[clip,width=1\hsize]{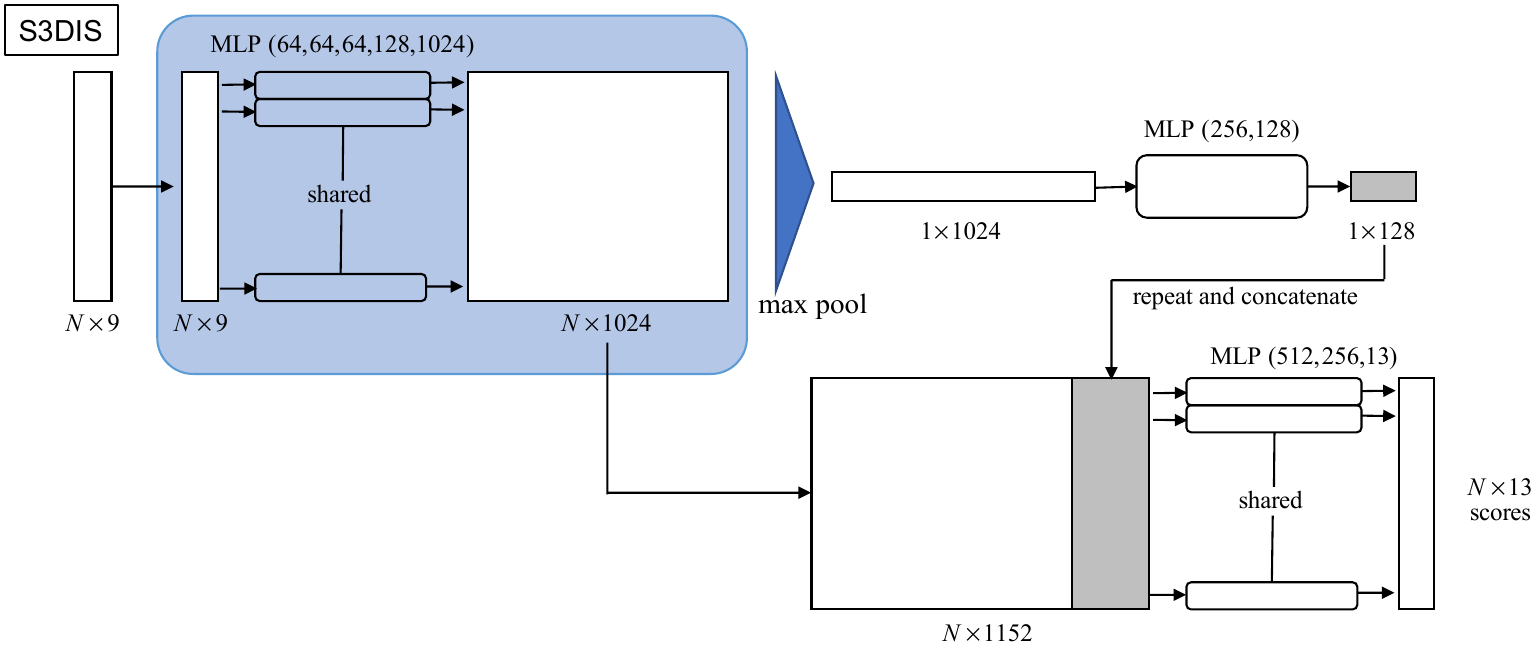}
    \end{tabular}
    \caption{PointNet architectures. The architecture design follows the author's implementation of PointNet. Colored regions indicate the embedding function, which is replaced with the Gaussian kernel for GPointNet. The numbers in the bracket of MLP indicate the output dimension of each layer; for example, MLP(256, 128) indicates the two-layer perceptron with 256 and 128 output units. The $N\times 16$ matrix of input to the ShapeNet model denotes one-hot category labels. Batch normalization and ReLU activation are used after all the layers except for the output layer}
    \label{fig:arc}
\end{figure}

We show the model architecture of PointNet in Fig. \ref{fig:arc}.
The architecture followed the author's implementation of PointNet.\footnote{https://github.com/charlesq34/pointnet.}
The architectures of GPointNet and LUTI-MLP are provided to replace the embedding function of PointNet (colored area in Fig. \ref{fig:arc}) with the Gaussian kernel and the table function.
The model for the part segmentation requires intermediate point-wise features, and both GPN w/Gaussian and LUTI-MLP use the extra Gaussian kernel and the table function $\hat{\phi}:\mathbb{I}^3\rightarrow\mathbb{R}^{832}$ to obtain the point-wise feature.

\begin{figure}[t]
    \centering
    \begin{tabular}{c}
    \includegraphics[clip,width=1\hsize]{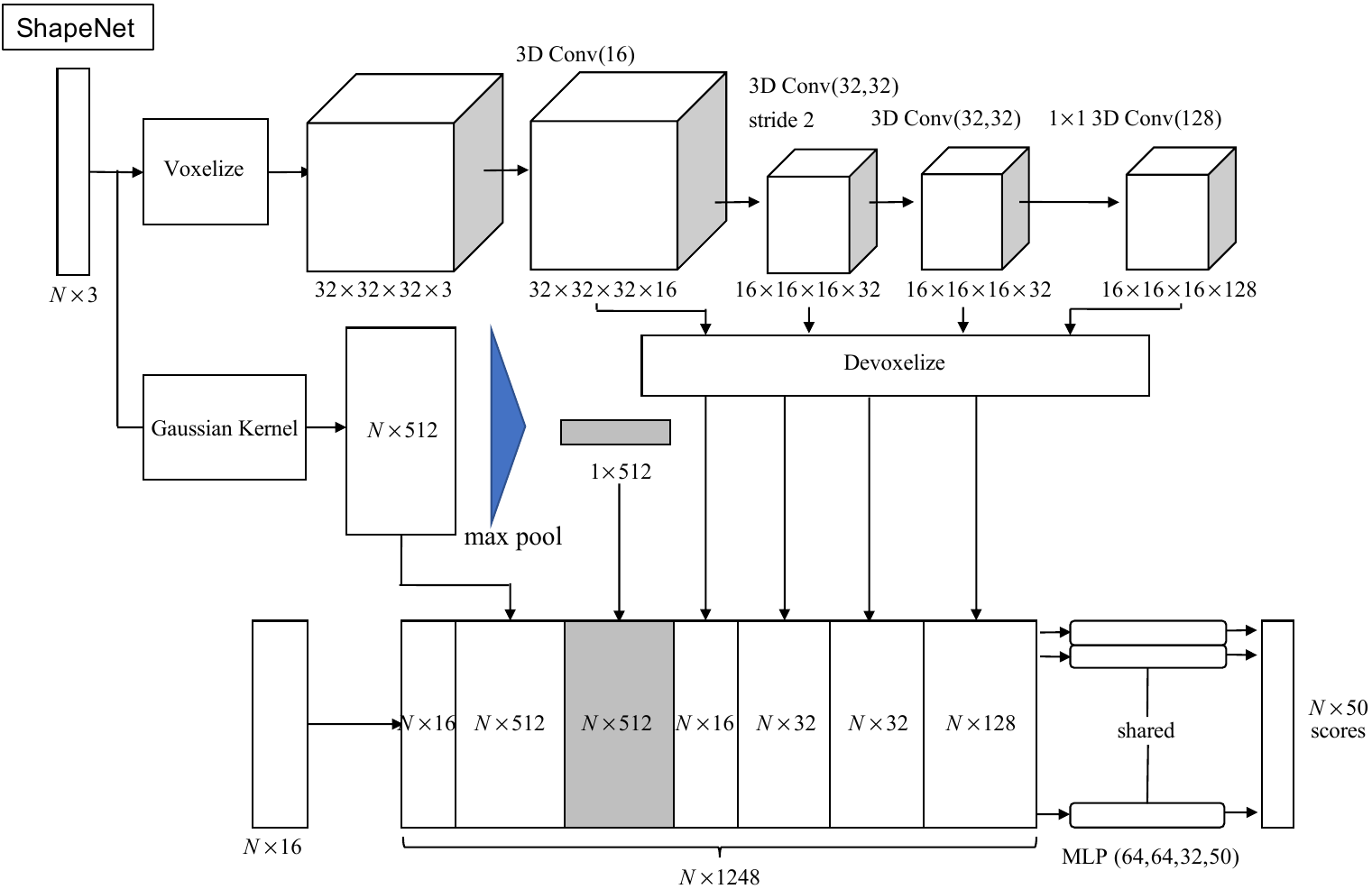}\\
    \includegraphics[clip,width=1\hsize]{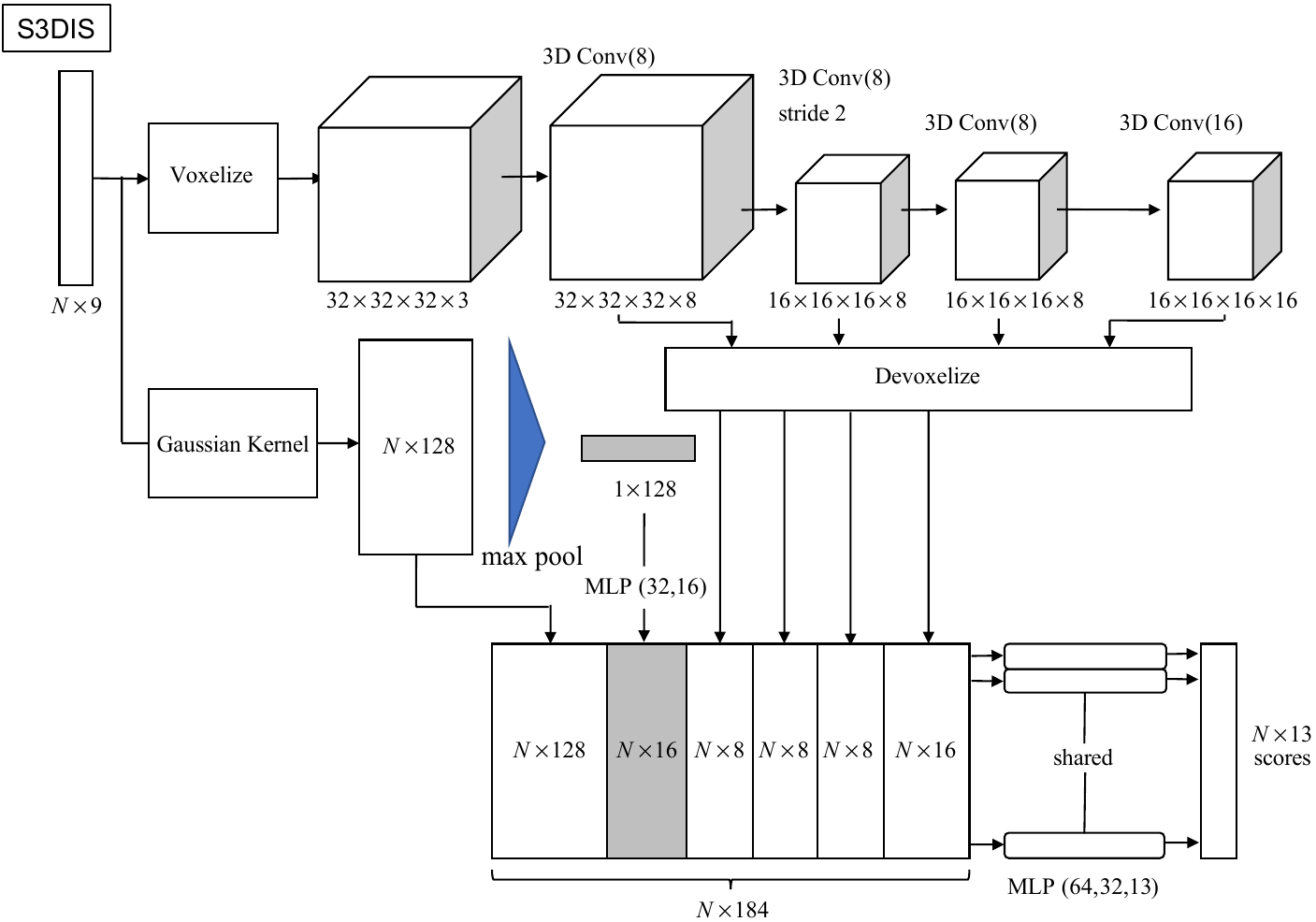}
    \end{tabular}
    \caption{PGN w/ Conv architectures. The numbers in the bracket of 3D Conv indicate the output channels of each volumetric convolution layer. The kernel size of all the volumetric convolution layers except for $1\times 1$ 3D Conv is 3, and that of the $1\times 1$ 3D Conv is 1. The $N\times 16$ matrix of input to the ShapeNet model denotes the one-hot category labels. Batch normalization and ReLU activation are used after all convolution layers}
    \label{fig:gpn_conv}
\end{figure}

For reproducing PVCNN, we utilize the author's implementation,\footnote{https://github.com/mit-han-lab/pvcnn.} and its architecture follows the implementation of \texttt{pvcnn/models/shapenet/pvcnn.py} and \texttt{pvcnn/models/s3dis/pvcnn.py}.
The architecture of GPN w/Conv is inspired by PVCNN, and we show the architectures in Figure \ref{fig:gpn_conv}.
The voxelization and devoxelization in Figure \ref{fig:gpn_conv} are the same as those used in PVCNN.

\section{Calculation of Floating-Point Operations}
\label{app:flops}
The matrix-vector product between $M\times N$ matrix and $N$-dimensional vector requires $2MN-M$ FLOPs, the vector-vector product between $N$-dimensional vectors requires $2N-1$ FLOPs, and the $N$-dimensional-vector-scalar product requires $N$ FLOPs~\cite{flops}.
The $k$-th Gaussian kernel consists of vector-vector subtraction, matrix-vector product, vector-vector product, and $\exp(\cdot)$ for a single point.
Thus, under $\boldsymbol{x}_i\in\mathbb{R}^M$, $\boldsymbol{\mu}_k\in\mathbb{R}^M$, and $\mathbf{\Sigma}_k\in\mathbb{R}^{M\times M}$, the $k$-th Gaussian kernel requires the following FLOPs:
\begin{align}
    \underbrace{M}_{\text{vector-vector sub.}}+\underbrace{2M^2-M}_{\text{matrix-vector prod.}}+\underbrace{2M-1}_{\text{vector-vector prod.}}+\underbrace{E}_{\exp(\cdot)},
\end{align}
and then the FLOPs/sample of the Gaussian kernel is $K\cdot N\cdot(2M^2+2M-1+E)$ because there are $K$ indicators and $N$ points.

PointNet consists of five-layer perceptron, and the intermediate layers do not depend on the input dimension $M$ and the output dimension $K$.
These intermediate layers require $2\cdot(2\cdot 64\cdot 64 - 64) + 2\cdot128\cdot64 - 128 = 32512$ FLOPs for a single point on the model for ModelNet40.
The input layer and the output layer require $2\cdot 64\cdot M-64$ FLOPs and $2\cdot 128\cdot K - K$ FLOPs, respectively.
Thus, FLOPs/sample of PointNet is given as $N\cdot(128M+255K+32448)$ FLOPs/sample.
Note that we ignore FLOPs of batch normalization and ReLU activation because these FLOPs are sufficiently few when compared with FLOPs of the matrix-vector product.

LUTI-MLP mainly consists of linear interpolation, which is described as the weighted sum of the vertexes of an $M$-dimensional cube.
The number of the vertexes is $2^M$, and the linear interpolation is defined as follows:
\begin{align}
    \sum_{\boldsymbol{v}\in\mathcal{V}(\boldsymbol{x}_i)}\prod_m^{M}w_m\boldsymbol{v},
\end{align}
where $\mathcal{V}(\boldsymbol{x}_i);|\mathcal{V}(\boldsymbol{x}_i)|=2^M$ denotes the set of vectors in the lookup table corresponding to $\boldsymbol{x}_i$.
Therefore, LUTI-MLP requires $2^M\cdot M$ times vector-scalar product and $2^M-1$ times vector sum.
Because of $\boldsymbol{v}\in\mathbb{R}^K$, the vector-scalar product requires $K\cdot 2^M\cdot M$, and the vector sum requires $K\cdot(2^M-1)$.
As a result, LUTI-MLP requires $K\cdot N\cdot(2^M\cdot M+2^M-1)$ FLOPs/sample.
Note that we ignore FLOPs of computing $w_m$.

PVCNN consists of point-based transformation and voxel-based transformation, and voxel-based transformation consists of voxelization, volumetric convolution, and devoxelization.
Because the volumetric convolution operation for a single voxel is defined as a matrix-vector product between the $c^\prime\times ck^3$ matrix and the $ck^3$-dim vector, FLOPs of the volumetric convolution is given by $r^3\cdot(2c^\prime\cdot c\cdot k^3-c')$, where $r$, $c$, $c^\prime$, and $k$ denote the spatial resolution, the number of input channels, the number of output channels, and the kernel size, respectively.
For example, since PVCNN (0.25$\times$ C) has two 3D convolution layers per block, the computational costs of the first layer of the ShapeNet model are given by $32^3\cdot(2\cdot16\cdot3\cdot3^3-16)+32^3\cdot(2\cdot16\cdot16\cdot3^3-16)=537$M FLOPs/sample.
FLOPs/sample of the other operations of PVCNN (\textit{e.g.}, point-based feature transformation and trilinear interpolation for devoxelization) are calculated by the same procedure as PointNet and LUTI-MLP.



\end{appendix}

\end{document}